\documentclass[conference]{IEEEtran}
\IEEEoverridecommandlockouts

\usepackage{cite}
\usepackage{amsmath,amssymb,amsfonts}
\usepackage{algorithmic}
\usepackage{graphicx}
\usepackage{textcomp}
\usepackage{xcolor}
\usepackage{booktabs}
\usepackage{longtable}
\usepackage{multicol}
\usepackage{hyperref}

\def\BibTeX{{\rm B\kern-.05em{\sc i\kern-.025em b}\kern-.08em
    T\kern-.1667em\lower.7ex\hbox{E}\kern-.125emX}}
\begin{document}

\title{Label-Guided Imputation via Forest-Based Proximities for Improved Time Series Classification
}

\author{
\IEEEauthorblockN{
Jake S. Rhodes\IEEEauthorrefmark{1}, 
Adam G. Rustad\IEEEauthorrefmark{2}, 
Sofia Pelagalli Maia\IEEEauthorrefmark{1}, 
Evan Thacker\IEEEauthorrefmark{3}, \\
Hyunmi Choi\IEEEauthorrefmark{4}, 
Jose Gutierrez\IEEEauthorrefmark{5}\IEEEauthorrefmark{6}, 
Tatjana Rundek\IEEEauthorrefmark{5}, 
Ben Shaw\IEEEauthorrefmark{7}}
\vspace{0.5em}
\IEEEauthorblockA{\IEEEauthorrefmark{1}Department of Statistics, Brigham Young University, Provo, UT, USA \\
Emails: rhodes@stat.byu.edu, smaia13@byu.edu}
\IEEEauthorblockA{\IEEEauthorrefmark{2}Department of Computer Science, Brigham Young University, Provo, UT, USA \\
Email: arusty@byu.edu}
\IEEEauthorblockA{\IEEEauthorrefmark{3}Department of Public Health, Brigham Young University, Provo, UT, USA \\
Email: elt@byu.edu}
\IEEEauthorblockA{\IEEEauthorrefmark{4}Department of Neurology, Vagelos College of Physicians \& Surgeons, Columbia University, New York, NY, USA \\
Email: hc323@cumc.columbia.edu}
\IEEEauthorblockA{\IEEEauthorrefmark{5}Department of Neurology, Miller School of Medicine, University of Miami, Miami, FL, USA \\
Email: trundek@med.miami.edu}
\IEEEauthorblockA{\IEEEauthorrefmark{6}Elmhurst Hospital Center, NYC Health + Hospitals, New York, NY, USA \\
Email: drjosegc@hotmail.com}
\IEEEauthorblockA{\IEEEauthorrefmark{7}Department of Mathematics \& Statistics, Utah State University, Logan, UT, USA \\
Email: ben.shaw@usu.edu}
}

\maketitle

\begin{abstract}
Missing data is a common problem in time series data. Most methods for imputation ignore label information pertaining to the time series even if that information exists. In this paper, we provide a framework for missing data imputation in the context of time series classification, where each time series is associated with a categorical label. We define a means of imputing missing values conditional upon labels, the method being guided by powerful, existing supervised models designed for high accuracy in this task. From each model, we extract a tree-based proximity measure from which imputation can be applied. We show that imputation using this method generally provides richer information leading to higher classification accuracies, despite the imputed values differing from the true values. 
\end{abstract}


\section{Introduction}

Missing data is a common and challenging issue in the preprocessing pipeline for machine learning, often impairing model performance and complicating analysis. This problem is prevalent in time series data, which frequently contains missing values due to a variety of factors, such as irregular sampling intervals, human error or data censoring, sensor or instrument malfunction, incomplete logging or delayed data reporting, and data corruption during transmission or storage~\cite{pratama2016review-of-missing}.

In this work, we focus on addressing the missing data problem within the context of time series classification, where each time series instance is associated with a categorical label, and the primary goal is to predict the label based on the observed values or derived features. Unlike forecasting that focuses on next-point prediction, our emphasis lies in improving classification performance in the presence of missing values.

For this purpose, we propose a new methodology that integrates supervised metric learning with tree-based ensemble models designed for time series classification. Specifically, we use random forest-derived proximity measures~\cite{breiman2001randomforests} to construct task-aware similarities between incomplete time series. These proximities, inspired by the geometry- and accuracy-preserving proximities (RF-GAP)~\cite{rhodes2023rfgap}, allow us to impute missing values as weighted averages over similar samples of non-missing data in the training set. This approach maintains alignment with the supervised learning objective, leading to informed imputations that respect both feature and label structure.

\section{Related Work}

We begin by reviewing relevant background on time series classification (Section~\ref{sec:ts-class}) and random forest proximities (Section~\ref{sec:gap-proximities}). These topics provide the necessary foundation for formally introducing our proximity-based imputation methods.

\subsection{Time Series Classification}\label{sec:ts-class}


Several machine learning tasks involve time series data. While time series forecasting is a common machine learning task, time series classification--the identification of entire time series with categorical labels--is another. While some time series classification models predict using the raw time series, many existing models rely on transforming the raw time series or extracting informative features before applying machine learning or deep learning models for classification~\cite{ye_shapelets_2009, hills_shapelettransform_2014, dempster_rocket_2020, dempster_multirocket_2023, christ_tsfresh_2018, middlehurst2022freshprince}. These transformations may include statistical summarization, frequency-domain representations, shapelets, or learned embeddings designed to capture discriminative characteristics of the series.


Time series classification methods can be broadly grouped into seven categories based on their primary function. \textbf{Convolution-based} approaches apply large sets of fixed or learned kernels to extract local temporal patterns without extensive preprocessing or feature engineering~\cite{dempster_rocket_2020, dempster_multirocket_2023}. \textbf{Dictionary-based} methods take inspiration from the bag-of-words model by converting sliding windows into symbolic words and representing series through word frequency histograms~\cite{schafer_weasel_2017, middlehurst_tde_2021}. In contrast, \textbf{distance-based} classifiers rely on elastic similarity measures, such as Dynamic Time Warping (DTW), to compare time series under temporal misalignments, often combined with $k$-NN or ensemble models~\cite{lucas_proximity_2019}. \textbf{Feature-based} approaches instead summarize series into fixed-length vectors using statistics, symbolic approximations, or signature transforms, enabling the application of conventional classifiers~\cite{fulcher_hctsa_2014, lubba_catch22_2019, christ_tsfresh_2018, kidger_signatures_2021}. Related to this, \textbf{interval-based} classifiers focus on extracting informative features from subseries, either randomly or with supervision, to capture localized temporal dynamics~\cite{deng_tsf_2013, lines_elastic_2015, middlehurst_drcif_2021, middlehurst_quant_2023, lubba_catch22_2019}. Another line of work centers on \textbf{shapelet-based} methods, which identify discriminative subsequences that serve as features or decision boundaries for classification~\cite{ye_shapelets_2009, hills_shapelettransform_2014, lemos_rsast_2022, faouzi_dilated_2023}. Finally, \textbf{deep learning} approaches leverage neural architectures such as the more recent InceptionTime family~\cite{Fawaz2020InceptionTime}.

Many of these methods, except for deep learning methods, are flexible in the underlying estimator used for the final classification step. In this work, we use methods that (1) natively use tree-based classifiers (decision trees, random forests~\cite{breiman2001randomforests}, and rotation forests~\cite{rodriguez2006rotation}, etc.), or (2) use a simple base estimator that can be easily replaced by a random forest for classification. The tree-based models allow us to directly construct measures of similarity between time series regardless of the feature space used to train the model, following the framework of~\cite{rhodes2023rfgap}. The time series models used in this paper are adapted from the AEON API~\cite {middlehurst2024aeon} to incorporate the tree-based similarities.

\subsection{Construction of Proximities}\label{sec:gap-proximities}

Random forests are supervised learning models composed of an ensemble of decision trees trained on bootstrap samples of the data~\cite{breiman2001randomforests}. Each tree partitions the feature space through recursive binary splits based on randomly selected subsets of features. This process results in terminal nodes (leaves) where similar observations are grouped together based on the model's learned decision boundaries.

These terminal nodes naturally create a means of defining similarity between points: the more frequently two points co-occur in the same terminal node across different trees, the more similar they are considered. This idea underlies the classic random forest proximity, computed as the fraction of trees in which two observations fall into the same leaf.

RF-GAP~\cite{rhodes2023rfgap} builds on this idea by incorporating information about terminal node size and bootstrapping to better capture how each training point influences the prediction, and are thus capable of perfectly reconstructing the random forest out-of-sample predictions as a weighted neighbor classifier.

For a training pair $(x_i, x_j)$, the RF-GAP proximity is defined as:

\begin{align}
p(i, j) = \frac{1}{|S_i|} \sum_{t \in S_i} \frac{I(x_j \in J_i(t)) \cdot c_j(t)}{|M_i(t)|}.
\end{align}

Here, $S_i$ is the set of trees in which $x_i$ is out-of-bag, $J_i(t)$ is the multiset of in-bag points that fall in the same terminal node as $x_i$ in tree $t$, $c_j(t)$ is the count of how many times $x_j$ appears in the in-bag sample for tree $t$, and $M_i(t)$ is the size of the multiset of in-bag points co-occurring with $x_i$ in that terminal node. RF-GAP weights the co-occurrence of a test point and training point in the same terminal node by the training point’s presence in the bootstrap sample and normalizes this by the node size.

Because test points are out-of-bag by definition, their similarity to training points can be computed using the same RF-GAP formula. This natural extension allows missing values in the test set to be imputed using supervised similarities that leverage class label information from the training set.

Time series models that predict using tree-based models can be adapted to construct GAP proximities by adjusting the tree-based mechanism used for the final prediction. For example, as RED CoMETS~\cite{luca2023redcomets} trains an ensemble of random forests on different representations learned from the Symbolic Aggregate approXimation (SAX)~\cite{lin2003sax} and Symbolic Fourier Approximation (SFA)~\cite{schaefer2012sfa} transformations, each forest can contribute a set of proximities that can be aggregated. TSFresh~\cite{christ_tsfresh_2018} and FreshPRINCEClassifier~\cite{middlehurst2022freshprince} each extract features from the time series data which are then used to train a rotation forest~\cite{rodriguez2006rotation} classification model.\footnote{It is important to note that although rotation forests do not use bagging in the same way that random forests do, they can be adapted to do this in order to calculate GAP proximities.}

Proximity Forests~\cite{lucas_proximity_2019} is a tree-based time series classification model that constructs an ensemble of decision trees using a diverse set of time series distance measures. At each node in a tree, a time series prototype is randomly selected along with a randomly chosen distance function—such as Dynamic Time Warping (DTW) variants, Edit Distance with Real Penalty (ERP), or Move-Split-Merge (MSM). The data points are then routed to child nodes based on their closest proximity to the prototype under the selected distance, allowing the forest to capture a variety of shape- and alignment-based similarities across time series. The authors of \cite{shaw2025forestproximitiestimeseries} introduced a similarity measure defined on proximity forests based on GAP proximities.


In this work, we selected time series classification models from various categories which are capable of using tree-based estimators. These models include FreshPRINCE (feature-based), Quant (interval-based), RDST (shapelet-based), ROCKET (convolution-based), and REDCoMETS (dictionary-based).

\section{GAP Imputation}

Let $\mathcal{D} = \{(x_n, y_n)\}_{n=1}^N$ denote a time series dataset of $N$ instances, where each instance $x_n \in \mathbb{R}^{p \times T}$ is a multivariate time series with $p$ features observed over $T$ time points, and $y_n \in \mathcal{Y}$ is the corresponding target label. For the purposes of this paper, $p = 1$ corresponds to univariate time series, although the methodology readily extends to $p > 1$ for multivariate settings for models capable of handling multivariate time series.

The full dataset is represented by a 3-dimensional array $X \in \mathbb{R}^{N \times p \times T}$, where the $(n, j, t)$-th entry, denoted $x_{njt}$, corresponds to the value of feature $j$ at time $t$ for instance $n$.

For each instance $n$ and feature $j \in \{1, \dots, p\}$, define the index sets of missing and observed time points:
\[
\mathcal{M}_{nj} = \{t : x_{njt} \text{ is missing} \}, \hspace{0.5em} 
\mathcal{O}_{nj} = \{t : x_{njt} \text{ is observed} \},
\]
so that $\mathcal{M}_{nj} \cup \mathcal{O}_{nj} = \{1, \dots, T\}$ for all $n, j$. Denote missing values as $x_{njt}^{\text{miss}}$ and observed values as $x_{njt}^{\text{obs}}$, with their imputations written as $\hat{x}_{njt}^{\text{miss}}$ and $\hat{x}_{njt}^{\text{obs}}$, respectively.

The GAP imputation procedure begins with an initial imputation of missing values in $X$ using basic strategies such as time-wise mean, median, or $k$-nearest neighbors ($k$-NN). This initialization can be performed globally across all time points or conditioned on the label $y_n$.

Once the imputed array, $\hat{X}$, is obtained, a time series classification model is trained on $(\hat{X}, y)$. From this model, GAP proximities $p(n, k)$ are derived for each pair of instances, quantifying similarity between point indices $n$ and $k$, as described in Section~\ref{sec:gap-proximities}. These proximities act as adaptive weights for imputation, linking missing entries to similar observed ones.

Imputation proceeds iteratively over features and time points. For a given missing entry $x_{njt}^{\text{miss}}$ with $t \in \mathcal{M}_{nj}$, the imputed value is computed using observed values at $s \in \mathcal{O}_{nj}$ and proximity weights. For continuous features, we use a proximity-weighted average:

\[
\hat{x}_{njt}^{\text{miss}} = \sum_{k : t \in \mathcal{O}_{kj}} p(n, k) \, x_{kjt}.
\]


For categorical features with class set $\mathcal{C}_j$, we use a weighted majority vote:

\[
\hat{x}_{njt}^{\text{miss}} = \arg\max_{c \in \mathcal{C}_j} \sum_{k : t \in \mathcal{O}_{nj}} p(n, k) \cdot \mathbf{1}(x_{kjt} = c).
\]

To internally validate the imputation process, the imputation is applied to observed values $x_{njt}^{\text{obs}}$, treating them as pseudo-missing. This enables internal performance monitoring, analogous to out-of-bag evaluation in random forests. After one full pass updating all entries (both missing and pseudo-missing), a new model is trained on the updated $\hat{X}$, proximities are recomputed, and the process is repeated. The algorithm runs for a fixed number of iterations (typically 5), and the imputation yielding the best internal metric is selected.

The selection criterion is based on the reconstruction quality of re-imputed observed entries $\hat{x}_{njt}^{\text{obs}}$, using metrics such as $R^2$ for continuous features or $F_1$ score for categorical features. Alternatives like RMSE, MAE, or accuracy may also be specified.

For test-set imputation, given a new time series dataset $X^{\text{test}} \in \mathbb{R}^{N_{\text{test}} \times p \times T_{\text{test}}}$, the same initialization procedure is applied (without conditioning on labels), followed by extension of the trained GAP model to compute proximities between training and test instances. We assume that test instances are aligned over the same $T_{\text{test}}$ time points, leaving the extension to variable-length sequences for future work. Each missing entry $x_{njt}^{\text{test}}$ is then imputed using observed training data in feature $j$, weighted by the proximities $p^{\text{test}}(n, k)$ between test and training instances. In this way, test-set imputations benefit from label-informed structure learned during training without requiring labeled test data.

\section{Common Time Series Imputation}

For univariate time series data, several methods can be applied to impute missing values. We compare our GAP-based imputation models to commonly used univariate time series imputation models. 
Comparison to deep learning-based imputation methods, as well as comparison to multivariate imputation methods, is reserved for future work.

Mean imputation is a simple approach where each missing value is replaced by the mean of the observed values across time. While easy to implement, this method can introduce bias and underestimate temporal variability. A more robust alternative is median imputation, which substitutes missing entries with the median of the observed values, offering better resilience to outliers \cite{acuna2004treatment}.

In $k$-nearest neighbors imputation, missing values are estimated based on the values at similar time points or across similar sequences, using a distance metric such as Euclidean distance or Dynamic Time Warping. For regularly sampled univariate time series, simple heuristics such as forward fill or backward fill are often used; these methods propagate the last observed value carried forward (LOCF) or the next observed value carried backward (NOCB), respectively~\cite{wright2021locf}. Although they maintain temporal continuity, they implicitly assume that values remain constant over time.

Interpolation techniques, including linear or spline interpolation, estimate missing values by fitting a continuous function through the observed points, and are well-suited for evenly spaced time series. 

Model-based imputation approaches leverage statistical models such as autoregressive models (AR), Kalman filters, or the Expectation-Maximization (EM) algorithm to estimate missing values based on temporal dependencies (e.g., seasonality and trend decomposition using Locally Estimated Scatterplot Smoothing (STL)~\cite{cleveland1990stl}) and probabilistic structure. These methods can offer greater accuracy but often require stronger modeling assumptions and increased computational cost.

\section{Experimental Setup}

In time series data, missing values can arise from a variety of operational, environmental, or systematic factors, and understanding the mechanism by which data are missing is critical for selecting appropriate imputation strategies and for interpreting results meaningfully~\cite{cheng2023comprehensive}. Three primary missingness mechanisms are typically considered. Missing Completely at Random (MCAR) describes situations where the probability of a missing value is independent of both observed and unobserved data. Missing at Random (MAR) arises when the probability of missingness depends only on observed values; for instance, heart rate data in a multivariate time series may be more likely to be missing during periods of low physical activity, as indicated by accelerometer readings~\cite{farrahi2019location}. In univariate time series, MAR can also occur when earlier values in the sequence influence the likelihood of subsequent missingness. Missing Not at Random (MNAR) occurs when the missingness depends on the value itself, even after conditioning on all observed data. This may happen in time series when patients withhold symptom reports during severe episodes or when sensors systematically fail under extreme conditions~\cite{Che2018rnn-missing}.

\section{Evaluation}

To evaluate the models, we use 85 datasets from the UCR Time Series Classification Archive~\cite{dau2018ucr}. Each dataset was first standardized, and we then removed values according to the three standard mechanisms, MCAR, MAR, and MNAR, at different levels of percentages missing: 5\%, 25\%, 50\%. All data removals, imputations, and classifications are repeated over three different random states. Since we are using univariate time series, MAR was simulated by introducing missing values conditional on the past values of the series. Specifically, if a lagged value (e.g., one time step earlier) exceeded a specified threshold, such as a given percentile, the corresponding current value was set to missing with a given probability (following the same percentages as above). This creates a scenario where the probability of missingness depends on previously observed data.  MNAR was simulated by introducing missing values based on the current value of the time series itself. If a value exceeded a predefined threshold, it was set to missing with a fixed probability. This setup captures the Missing Not At Random mechanism, where the likelihood of missingness depends directly on the unobserved value, making it systematically related to the data itself.

To evaluate the methods, we conducted two tasks: (1) Direct value comparison, in which we computed the root mean squared error (RMSE) between the true and imputed values; and (2) Post-imputation evaluation, where we assessed the classification accuracy on a test set after training models on data with imputed values.

\section{Results}

\begin{figure}[!htb]
    \centering
    \includegraphics[width=0.9\linewidth]{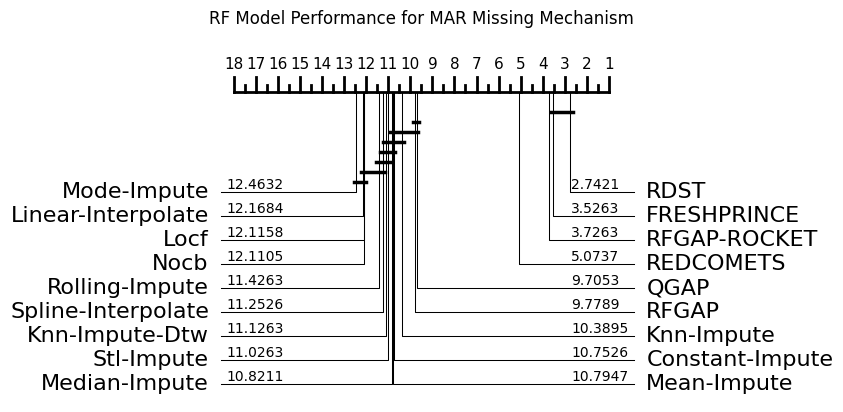}\\[1ex]
    \includegraphics[width=0.9\linewidth]{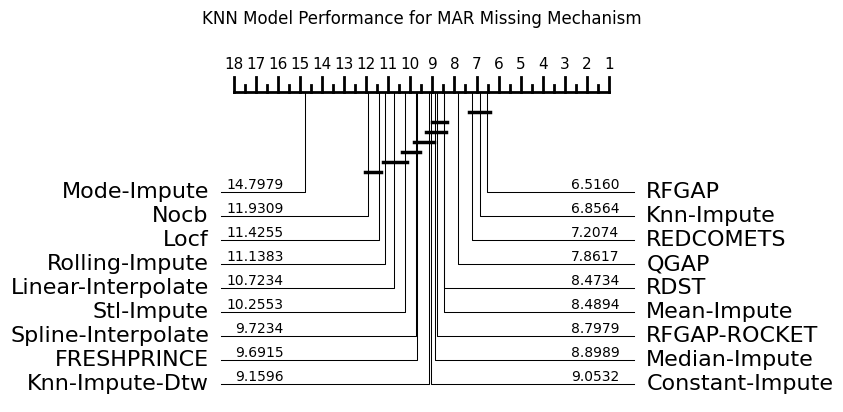}
    \caption{The critical difference plots using a random forest (top) or $k$-NN (bottom) classifier after imputation. In this example, the results are aggregated across each proportion of missingness under the MAR mechanism, although additional examples can be found on the code repository. Here we see that the random forest classifier favors the imputed data using the GAP-based models, with RDST, FreshPrince, and RF-GAP with ROCKET having the best ranks. For the $k$-NN model, RF-GAP, $k$-nn imputation, and REDCoMETS were the highest ranked. In each case, at least four of the five highest rankings are using the GAP-based imputation.}
    \label{fig:cd_mar}
\end{figure}

We begin by evaluating how accurately each imputation method recovers the original data, measuring the root mean squared error (RMSE) between the imputed values and the ground truth values that were artificially removed. This metric quantifies the point-wise consistency of each method in reconstructing the original data distribution.

Across the 85 standardized univariate datasets from UCR~\cite{dau2018ucr}, $k$-nearest neighbors ($k$-NN) imputation achieved the lowest average RMSE, indicating the most accurate recovery of missing values. The next best performers were REDCoMETS and RF-GAP, which also produced relatively low RMSE values. Most other imputation approaches, including DTW-based $k$-NN, QGAP, RDST, FreshPRINCE, ROCKET, and several simple statistical techniques (e.g., median, mean, linear, and rolling interpolation), exhibited higher RMSE. Mode imputation was excluded from the comparison because its RMSE exceeded 1.0 and was considered an outlier. Overall, the ranking of methods was consistent across different missingness mechanisms and percentages of missing data, with $ k$-NN.

We note that lower reconstruction error does not always translate directly into better downstream learning performance~\cite{le2021whatsagood}. Since the primary goal of imputation in our setting is to support classification under missingness, we assess how different imputation methods impact predictive accuracy. To this end, we train random forest and $k$-NN classifiers on the imputed datasets and evaluate their test accuracy. The results under the MAR mechanism are found in Figure~\ref{fig:cd_mar} (similar results were found using other mechanisms). For the random forest model, GAP-based approaches consistently achieve the highest ranks, outperforming the other imputation strategies. For the $k$-NN classifiers, RF-GAP has the highest ranking, led by $k$-NN imputation and REDCoMETS. In each case, at least four of the top five methods involve GAP-based imputation. A full set of results can be viewed in the code repository at \href{https://github.com/JakeSRhodesLab/TS-Supervised-Imputation}{https://github.com/JakeSRhodesLab/TS-Supervised-Imputation}.

\section{Case Study}

The Northern Manhattan Study (NOMAS) is a longitudinal cohort study conducted in the multi-ethnic community of northern Manhattan, New York City, which included 3,497 adults who were 40 years of age or older when they enrolled in the study between 1993 and 2001~\cite{sacco1998stroke}. 

As part of cohort follow-up, a neuropsychological battery was administered in English or Spanish, according to participant preference, which included measures of episodic memory, processing speed, semantic memory, and executive function~\cite{gardener2016ideal}.

In our study, five measured cognitive items were used to assess imputation and subsequent classification. The cognitive items used and missingness percentages are as follows: Letter Fluency Test (43.1\% missing), Color Trails 1 (42.6\% missing), Color Trails 2 (45.4\% missing), Grooved Pegboard (42.8\% missing), and Symbol Digit Modalities Test (62.9\% missing).

We analyzed longitudinal cognitive data from 1,266 participants who had between 1 to 10 measures obtained over a 28-year follow-up period. Data with less than 10 measures were padded with missing values prior to imputation. Incident dementia was used as the response in the model. Participants with incident dementia were identified by a team of neuropsychologists and neurologists using diagnostic criteria from the National Institute on Aging—Alzheimer’s Association (NIA-AA) and the Diagnostic and Statistical Manual of Mental Disorders, 5th edition (DSM-5), including evidence of decline in cognitive performance over time, functional impairment, and lack of psychiatric or other diagnosis that would explain the cognitive dysfunction~\cite{wright2021locf}.

Since no ground truth values are available---some measurements were never taken at certain time steps---we evaluate time series imputation indirectly through post-imputation classification performance, using random forest and $k$-NN models. These models were trained on imputed training variables, where each variable was treated as a univariate time series and imputed individually. The same imputation models were then applied to a held-out test set. For unsupervised methods, the imputation was applied directly to the test set. For supervised models, we used the RF-GAP direct extension to the test set. Classification results were averaged over three random seeds, and are presented in Table~\ref{tab:nomas-accuracies}.

\begin{table}[!htb]
\centering
\caption{The $k$-NN and random forest model performance summary (mean $\pm$ std.) across the described NOMAS variables after applying the imputation methods. The scores are sorted in descending order by $k$-NN accuracy. The top model is bold, the second best being underlined. The imputation based on the time series classification (marked with an asterisk) generally provided more sufficient information for post-imputation classification, yielding higher accuracies for both classification models, though the differences are more marked in the $k$-NN case.}
\label{tab:nomas-accuracies}
\begin{tabular}{lccc}
\toprule
Model & KNN Accuracy & RF Accuracy \\
\midrule
REDCoMETS* & \textbf{0.848 $\pm$ 0.00} & \textbf{0.848 $\pm$ 0.00} \\
FreshPRINCE* & \underline{0.847 $\pm$ 0.00} & \underline{0.847 $\pm$ 0.00} \\
RFGAP* & \underline{0.847 $\pm$ 0.00} & \underline{0.847 $\pm$ 0.00} \\
RFGAP Rocket* & 0.845 $\pm$ 0.01 & 0.842 $\pm$ 0.01 \\
QGAP* & 0.843 $\pm$ 0.01 & 0.830 $\pm$ 0.04 \\
RDST* & 0.840 $\pm$ 0.02 & 0.843 $\pm$ 0.01 \\
STL Impute & 0.838 $\pm$ 0.01 & 0.833 $\pm$ 0.01 \\
Median Impute & 0.838 $\pm$ 0.00 & 0.846 $\pm$ 0.00 \\
Mode Impute & 0.835 $\pm$ 0.02 & 0.796 $\pm$ 0.09 \\
KNN Impute (DTW) & 0.832 $\pm$ 0.01 & \underline{0.847 $\pm$ 0.00} \\
NOCB & 0.830 $\pm$ 0.01 & 0.841 $\pm$ 0.01 \\
KNN Impute & 0.827 $\pm$ 0.02 & 0.842 $\pm$ 0.01 \\
Spline Interpolate & 0.826 $\pm$ 0.02 & 0.844 $\pm$ 0.01 \\
Constant Impute & 0.826 $\pm$ 0.02 & 0.842 $\pm$ 0.02 \\
LOCF & 0.826 $\pm$ 0.02 & 0.845 $\pm$ 0.00 \\
Mean Impute & 0.825 $\pm$ 0.03 & \underline{0.847 $\pm$ 0.00} \\
Linear Interpolate & 0.809 $\pm$ 0.06 & 0.790 $\pm$ 0.07 \\
Rolling Impute & 0.783 $\pm$ 0.10 & 0.709 $\pm$ 0.22 \\
\bottomrule
\end{tabular}
\end{table}

When using the random forest classifier, performance is relatively consistent across different imputation methods, suggesting that the random forest can effectively learn from most imputation strategies, although the supervised GAP methods (marked with an asterisk) generally achieve higher accuracy. In contrast, when using the $k$-NN classifier, the GAP methods yield substantially better performance than classical imputation methods. This suggests that the supervised imputation enhances the structure of the data in a way that benefits classification, such that a typically weaker model like $k$-NN can achieve performance comparable to the stronger random forest model when trained on data imputed with supervised methods.

\section{Conclusion}

In this paper, we introduced a framework for imputing missing values in time series classification tasks by leveraging label-guided proximities derived from supervised tree-based models. Unlike traditional imputation approaches that operate independently of the classification objective, our method uses proximity measures that reflect both the feature and label structure of the data. By incorporating this task-aware similarity into the imputation process, we provide a mechanism that aligns data reconstruction with downstream predictive goals.

Our empirical evaluations across 85 datasets from the UCR Time Series Archive demonstrate that, while traditional methods such as $k$-nearest neighbors can achieve lower reconstruction error (RMSE), GAP-based imputation models typically deliver better classification accuracy post-imputation, highlighting an advantage of imputing missing data using information from the supervised task. 

We applied our methods in a case study on cognitive decline in the NOMAS cohort. In this study, we showed that even simple classifiers like $k$-NN benefit from supervised imputation, bringing the classification accuracy to a level of more advanced random forest models. The results suggest that our approach is suited to domains where accurate classification is prioritized over exact value recovery, and where label information is available during training. The framework is applicable to multivariate classification tasks, but our current study is limited to univariate time series. Future work will extend the evaluation to multivariate and irregularly sampled series.

\section{Acknowledgments}

Funding for support and use of NOMAS data is provided by the following grants from the National Institutes of Health: R01AG074355 (Choi, Gutierrez, Thacker), R01AG057709 (Guttierez), R01AG066162 (Guttierez), and R56NS029993 (Rundek).

\bibliographystyle{IEEEtran}
\bibliography{main}

\end{document}